# On the Effects of Idiotypic Interactions for Recommendation Communities in Artificial Immune Systems

**Steve Cayzer**

Hewlett-Packard Laboratories

Filton Road

Bristol

BS12 6QZ

Steve_Cayzer@hp.com

**Uwe Aickelin**

School of Computer Science

University of Nottingham

NG8 1BB   UK

uxa@cs.nott.ac.uk

## Abstract

It has previously been shown that a recommender based on immune system idiotypic principles can outperform one based on correlation alone. This paper reports the results of work in progress, where we undertake some investigations into the nature of this beneficial effect. The initial findings are that the immune system recommender tends to produce different neighbourhoods, and that the superior performance of this recommender is due partly to the different neighbourhoods, and partly to the way that the idiotypic effect is used to weight each neighbour's recommendations.

## 1   INTRODUCTION

The idiotypic effect builds on the premise that antibodies can match other antibodies as well as antigens. It was first proposed by Jerne [6] and formalised into a model by Farmer et al [3]. The theory is currently debated by immunologists, with no clear consensus yet on its effects in the humoral immune system [5]. In a previous paper [1], we have shown that the incorporation of idiotypic effects can be beneficial for Artificial Immune System based recommender systems.

However, in that paper we did not explore the mechanisms of that beneficial effect. Such an exploration would seem worthwhile, particularly if this results in identifying the underlying causes of the improvements of the 'characteristics' of a community (either by changing its membership, or by evaluating the relative merit of each member). Such an effect will be generally useful in a range of applications, of which recommender systems provide just one example. In addition, a deeper understanding of the idiotypic effect may prove useful to the designers of other Artificial Immune System applications.

In this paper, we present the results of work undertaken to better understand the idiotypic effect. In order to set the context, the next section provides a definition of the idiotypic effect and the following one a brief review of Artificial Immune System based recommenders. We then present and discuss the results of our analysis to date.

## 2   IDIOTYPIC EFFECTS

The idiotypic network hypothesis was first proposed by Jerne [6]. It builds on the recognition that antibodies can match other antibodies as well as antigens. Hence, an antibody may be matched by other antibodies, which in turn may be matched by yet other antibodies. This activation can continue to spread through the population. The idiotypic network has been formalised by a number of theoretical immunologists in [7]. This theory could help explain how the memory of past infections is maintained. Furthermore, it could result in the suppression of similar antibodies thus encouraging diversity in the antibody pool.

The following is a formal equation for the idiotypic effect adapted from Equation 3 from Farmer [3]:

$$\begin{aligned}\frac{dx_i}{dt} &= c\left[\begin{pmatrix}\text{antibodies}\\\text{recognised}\end{pmatrix} - \begin{pmatrix}I\ am\\\text{recognised}\end{pmatrix} + \begin{pmatrix}\text{antigens}\\\text{recognised}\end{pmatrix}\right] - \begin{pmatrix}\text{death}\\\text{rate}\end{pmatrix}\\ &= c\left[\sum_{j=1}^{N} m_{ji} x_i x_j - k_1 \sum_{j=1}^{N} m_{ij} x_i x_j + \sum_{j=1}^{n} m_{ji} x_i y_j\right] - k_2 x_i \quad (1)\end{aligned}$$

Where:

$N$ is the number of antibodies

$n$ is the number of antigens.

$x_i$ (or $x_j$) is the concentration of antibody $i$ (or $j$)

$y_j$ is the concentration of antigen $j$

$c$ is a rate constant

$k_1$ is a suppressive effect and $k_2$ is the death rate

$m_{ji}$ is the matching function between antibody $i$ and antibody (or antigen) $j$

As can be seen from the above equation, the nature of an idiotypic interaction can be either positive or negative. Moreover, if the matching function is symmetric, then the balance between "I am recognised" and "Antibodies recognised" (parameters $c$ and $k_1$ in the equation) wholly determines whether the idiotypic effect is positive or negative, and we can simplify the equation. We can simplify the equation still further if we only allow one antigen in the Artificial Immune System. The simplified equation looks like this:

$$\frac{dx_i}{dt} = k_1 m_i x_i y - \frac{k_2}{n} \sum_{j=1}^{n} m_{ij} x_i x_j - k_3 x_i \quad (2)$$

Where:

$k_1$ is stimulation, $k_2$ suppression and $k_3$ death rate

$m_i$ is the correlation between antibody $i$ and the (sole) antigen

$x_i$ (or $x_j$) is the concentration of antibody $i$ (or $j$)

$y$ is the concentration of the (sole) antigen

$m_{ij}$ is the correlation between antibodies $i$ and $j$

$n$ is the number of antibodies.

## 3 RECOMMENDER SYSTEM

At this point, it is worth reviewing how this model can be applied to recommender systems. Full details can be found in [1], but a brief overview follows.

Recommender systems are those that use collaborative filtering techniques to produce predictions and recommendations [4]. So for example a movie recommender system would, given a film, provide a *prediction* for that film (i.e. an estimated rating for you). It might also provide a list of *recommended* films (i.e. films which it estimates that you would prefer over others). It does this by comparing users together (based on their votes for movies), and preparing some 'neighbourhood' of like-minded users from which it can produce predictions and recommendations.

The main loop of the recommender algorithm is shown in Figure 1 and is the core of our Artificial Immune System. The aim of this algorithm is to increase the concentrations of those antibodies (database users) that are similar to the antigen (target user) and yet different from each other. The process is thus subject to the suppression of similar antibodies following Jerne's idiotypic ideas mentioned above. Thus, over time the Artificial Immune System contains high concentrations of a diverse set of users who have similar film preferences to the target user.

The algorithm is terminated either when there are no more users to try, or when the Artificial Immune System is *stabilised*, i.e. it is full, and has not changed in consistency for more than ten iterations. The concentrations and correlations of the users in the final neighbourhood, i.e. final immune system iteration, are then used to calculate a weighted sum of the ratings of movies.

```
Initialise Artificial Immune System
Encode user for whom to make predictions as
antigen Ag
WHILE (Artificial Immune System not stabilised)
& (More data available) DO
  Add next user as an antibody Ab
  Calculate matching score between Ab and Ag
  Calculate matching scores between Ab and other
antibodies
WHILE (Artificial Immune System at full size) &
(Artificial Immune System not stable) DO
      Iterate Artificial Immune System
  OD
OD
```

Figure 1: Main loop of the Artificial Immune System's algorithm for recommendation.

Our previous work [1] compared two predictors, one based on a Simple Pearson test and one on our Artificial Immune System. In each case, a test user is taken from a database, and then predictions and recommendations are made for that user. Both predictors work by finding a neighbourhood and using that neighbourhood to produce predictions and recommendations.

Prediction quality is assessed by measuring the mean absolute error (details in [1]). Recommendation quality is assessed by comparing the ranked recommendations with the user's ranked ratings for the recommended films. Kendall's Tau can now be applied. This measure reflects the level of concordance in the lists, and proceeds by counting the number of discordant pairs. To do this we order the films by actual vote and apply the following formulae to the recommended films:

$$\tau = 1 - \frac{4N_D}{n(n-1)}$$

$$N_D = \sum_{i=1}^{n} \sum_{j=i+1}^{n} D(r_i, r_j) \quad (3)$$

$$D(r_i, r_j) = \begin{cases} 1 \text{ if } r_i > r_j \\ 0 \text{ otherwise} \end{cases}$$

Where:
$n$ is the overlap size
$r_i$ is the actual rank of film $i$ as recommended by the neighbourhood.

Note that $i$ here refers to the recommended rank of the film, not the film ID. $N_D$ is the number of discordant pairs, or, equivalently, the expected cost of a bubble sort to reconcile the two lists. $D$ is set to one if the rankings are discordant.

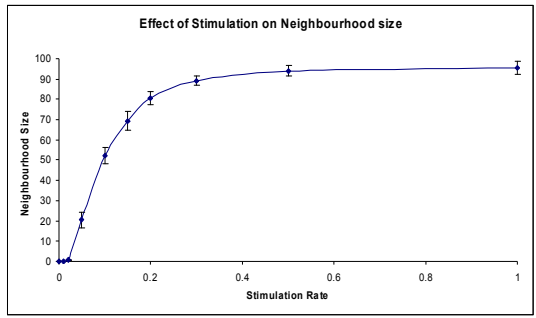 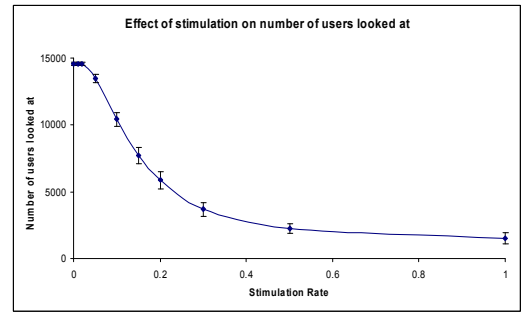

Figure 2: Effect of stimulation rate on neighbourhood size and reviewers looked at.

For the Simple Pearson case, the neighbourhood is composed of the 'top N' correlated users, where correlation is measured by the Simple Pearson statistical measure. In the Artificial Immune System case, the neighbourhood is created by building an immune system with the test user as the antigen, the neighbours as antibodies, and the Simple Pearson measure as a matching function. (In fact, in our experiments, this measure was weighted by the a fraction proportional to the number of films both users had seen, in order to penalise correlations made on the basis of only a few films). The behaviour of the neighbourhood is then governed by equation 2, with poorly performing antibodies being deleted from the neighbourhood. Note that we have treated the idiotypic effect as suppressive.

## 4   ANALYSIS OF EFFECTS

Although both the Artificial Immune System and Simple Pearson recommender algorithms are based on Pearson correlations, they act differently for a number of reasons:

- The choice of neighbours is different. In the Simple Pearson, the 100 highest correlated users (or all users that show any correlation, if this is less than 100) are chosen to form a neighbourhood. In the Artificial Immune System, this general rule is followed, except that stimulation adds threshold and idiotypic effect adds diversity.

- Even given the same neighbours, the weighting is different. In the Simple Pearson, the neighbour weight is simply the correlation between that neighbour and the test user. In the Artificial Immune System, this correlation is multiplied by that antibody's (neighbour's) concentration, which in turn is determined by running the Artificial Immune System algorithm over the neighbourhood.

To deal with the first point, the stimulation rate provides some fixed threshold for the correlation of any antibody with the antigen. Even in the absence of any idiotypic interactions, an antibody's correlation (weighted by the stimulation rate) must outweigh the death rate; otherwise, it will not survive in the Artificial Immune System. So, at low stimulation rates it may prove difficult to fill the Artificial Immune System completely. Conversely, at very high stimulation rates it may not be necessary to examine all the supplied users in order to fill an Artificial Immune System.

This effect was noted in our previous paper [1] and can be seen in Figure 2. Such a thresholding effect has been shown to be beneficial by Gokhale [4] in maintaining the quality of a neighbourhood by filtering out poorly correlated users (the Simple Pearson will consider all reviewers who have at least one vote in common with the test user).

Thus, the idiotypic effect should be viewed in the context of providing further refinement to a neighbourhood that is already known to be in some sense 'good'. Since the effect (in our model) is always negative, its impact may be to improve diversity by removing 'suboptimal' users from the Artificial Immune System. Conversely, it might be that the idiotypic effect is effective because, given a neighbourhood, it changes the weight of each neighbour (or concentration of each antibody) in that neighbourhood. This is the second point highlighted above.

In order to test out these hypotheses, we took a sample result, based on 100 predictions for detailed analysis. The 3 settings for each algorithm were as detailed in [1] except that default votes were not used. Thus, if a neighbour has not seen a film then that neighbour is ignored when making a prediction for that film. The Artificial Immune System parameters were set to 'good' values (as observed in the previous paper): thus stimulation rate was set to 0.3 and suppression rate to 0.2. As reported previously, the prediction performance (mean absolute error) was not significantly different between the two algorithms, but recommendation (Kendall's Tau) was significantly better for the Artificial Immune System recommender (as before, a Wilcoxon matched pairs signed rank test was used to assess significance).

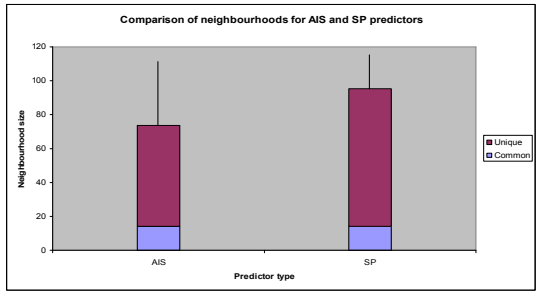

Figure 3: Comparison of Artificial Immune System and Simple Pearson neighbourhoods. The total size of each bar represents the total size of the neighbourhoods produced by each predictor (averaged over 100 predictions; bar shows standard deviation). The lower part of each bar shows the average number of common neighbours (i.e. appearing in both neighbourhoods). The remainder of the bar is composed of unique neighbours – that is, neighbours who appeared in one neighbourhood but not the other.

The first thing to observe is that the neighbourhoods produced by each algorithm are different. As implied from the above, Simple Pearson tended to produce large neighbourhoods (average 95.4 as opposed to 73.8 using the Artificial Immune System) and Figure 3 shows that the composition of these neighbourhoods is different. In particular, it does not seem that the Artificial Immune System neighbourhoods are merely subsets of the Simple Pearson neighbourhoods. In fact, the vast majority of neighbours are 'unique' – that is, chosen by one algorithm but not the other

Is it the neighbourhoods that make the difference to prediction and recommendation performance? Figure 4 shows Artificial Immune System and Simple Pearson performance on both neighbourhoods. For this experiment, we recorded the neighbourhoods found by both the Artificial Immune System and Simple Pearson algorithms.

We then reran the predictions, with everything the same except that this time we forced the Artificial Immune System and Simple Pearson algorithms to use our 'fixed' neighbourhoods. We can see that for prediction, changing the neighbourhood (or indeed algorithm) did not seem to make any significant difference (Table 1 has the details of the statistical tests). However, for recommendation, although the means are very similar (Fig 4), the Artificial Immune System neighbourhood usually produced better recommendations than the Simple Pearson neighbourhood (Table 1b). In fact, the neighbourhood effect seems to dominate, since given the Artificial Immune System neighbourhood, the Simple Pearson algorithm appears to do significantly better than the Artificial Immune System algorithm for recommendation. There is one exception to this trend, where the Artificial Immune System algorithm does not do significantly better for either neighbourhood. In addition, the Artificial Immune System algorithm does better on the Simple Pearson neighbourhood than the Simple Pearson algorithm does, indicating that the neighbour weightings, as well as the neighbours themselves, also contribute to the recommendation quality.

We ran these experiments using default votes (neighbours who had not voted on a film were assumed to give the film a slightly negative rating) and obtained similar results.

It is worth pointing out at this stage that these results should not be taken to be exhaustive, merely indicative. Indeed, we would not want to draw any firm conclusions based on only 100 predictions. This point will be returned to in the discussion. Nevertheless, the results obtained so far seemed to indicate that it was worth investigating the contribution of neighbourhood composition to recommendation performance.

**Fig 4a**                                        **Fig 4b**

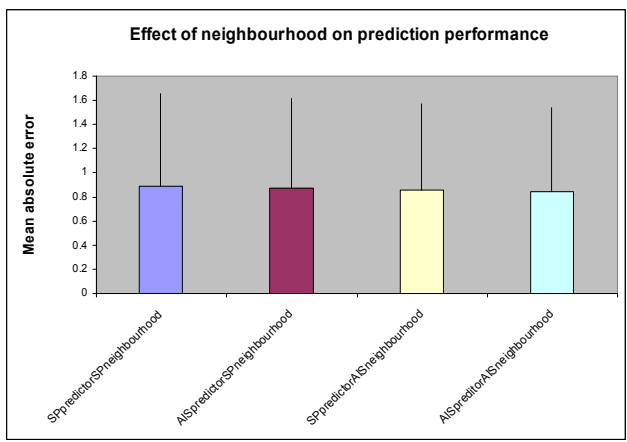 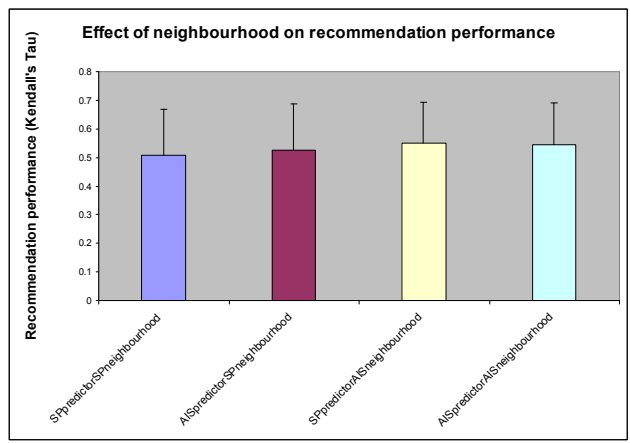

Figure 4: Effect of neighbourhood composition for Artificial Immune System and Simple Pearson algorithms. See text for details on fixing the neighbourhoods. Fig 4a shows prediction performance (measured as mean absolute error averaged over 100 predictions) for each algorithm and each neighbourhood. Fig 4b shows recommendation performance deviation. (measured as Kendall's Tau averaged over 100 predictions) for each algorithm and each neighbourhood. Bars show standard deviation.

Table 1: Analysis of differences between neighbourhoods and algorithms for both prediction (1a) and recommendation (1b). In each case, the Wilcoxon significance test was applied to the results obtained from each pair of regimes. Regimes that are significantly better are shown in **bold** (there were no significant differences found for prediction). [AIS = Artificial Immune System; SP = Simple Pearson]

**Table 1a**

| 1st Predictor | 1st neighbourhood | 2nd Predictor | 2nd neighbourhood | Median 1 | Median 2 | Number of (unequal) predictions compared | 1st regime better (sum of ranks) | 2nd regime better (sum of ranks) | Significance (upper bound) |
|---|---|---|---|---|---|---|---|---|---|
| SP | SP | AIS | SP | 0.682 | 0.697 | 97 | 2212 | 2541 | 0.5551 |
| SP | SP | SP | AIS | 0.682 | 0.658 | 97 | 2163 | 2590 | 0.4434 |
| SP | SP | AIS | AIS | 0.682 | 0.652 | 97 | 2176 | 2577 | 0.4717 |
| AIS | SP | SP | AIS | 0.697 | 0.658 | 97 | 2256 | 2497 | 0.6659 |
| AIS | SP | AIS | AIS | 0.697 | 0.652 | 97 | 2258 | 2495 | 0.6711 |
| SP | AIS | AIS | AIS | 0.658 | 0.652 | 84 | 1706 | 1864 | 0.7263 |

**Table 1b**

| 1st Predictor | 1st neighbourhood | 2nd Predictor | 2nd neighbourhood | Median 1 | Median 2 | Number of (unequal) predictions compared | 1st regime better (sum of ranks) | 2nd regime better (sum of ranks) | Significance (upper bound) |
|---|---|---|---|---|---|---|---|---|---|
| SP | SP | **AIS** | **SP** | 0.525 | 0.557 | 83 | 801 | 2685 | 1.917e-05 |
| SP | SP | **SP** | **AIS** | 0.525 | 0.549 | 83 | 707.50 | 2778.50 | 2.617e-06 |
| SP | SP | **AIS** | **AIS** | 0.525 | 0.542 | 85 | 930 | 2725 | 8.483e-05 |
| AIS | SP | **SP** | **AIS** | 0.557 | 0.549 | 82 | 1218.50 | 2184.50 | 0.02571 |
| AIS | SP | AIS | AIS | 0.557 | 0.542 | 80 | 1426 | 1814 | 0.3534 |
| **SP** | **AIS** | AIS | AIS | 0.549 | 0.542 | 78 | 2149 | 932 | 0.002459 |

We looked at a variety of neighbourhood parameters (we might term these community characteristics) across Simple Pearson and Artificial Immune System neighbourhoods. Four characteristics are of particular interest, and each will be discussed in turn. Firstly, it might seem reasonable to assume that performance improves with the number of neighbours in a neighbourhood. However, clearly there is a cost in collecting neighbours (of appropriate quality) together, and thus it will be useful if we can provide good quality recommendations from smaller neighbourhoods.

Another characteristic is the overlap size, which governs the number of recommendations we can assess (An overlap is a test user vote that is also contained in the union of all neighbours' votes). Thirdly, we looked at correlation between each neighbour and the test user. A high correlation shows that neighbours are clustered 'tightly' around the test user, which we might imagine would provide for better recommendations. Fourthly, the idiotypic effect is expected to reduce the inter-neighbour correlations. An obvious intuition might be that such a reduction causes an increase in recommendation quality.

Table 2 shows the difference in these community characteristics across Simple Pearson and Artificial Immune System neighbourhoods. It can be seen that the Artificial Immune System does produce neighbourhoods that are measurably different in character to the Simple Pearson neighbourhoods. In summary, the Artificial Immune System neighbourhoods are smaller, have less overlap, are generally less correlated with the test user and have lower inter-neighbour correlations.

In order to test out which (if any) of these characteristics is crucial, we plotted recommendation performance against each for the Artificial Immune System algorithm. The results seem to show that none of these characteristics *on their own* influences the performance in a clear way. Figure 5 shows scatter plots generated for each characteristic against recommendation quality. Trend lines (based on a power law) have been added to emphasise any underlying data trends.

The first plot suggests that neighbourhood size is not essential in order to obtain high quality recommendations. The second plot, however, does suggest that small overlap sizes might be beneficial for producing good recommendations (regression analysis has not been performed so at this stage this is merely a suggestion). This in some sense is intuitive, as it might be easier to produce higher quality recommendations if there are less of them. However, a balance needs to be struck here; once the overlap size gets too low, the neighbourhood may no longer prove useful to the user.

The third plot shows that, perhaps surprisingly, high correlation between neighbours and the test user may not be essential for high quality recommendations. Finally, the fourth plot would seem to indicate that reduced inter-neighbour correlation is not important in recommendation accuracy, or at least if it is responsible, it is part of a wider effect.

Table 2: Analysis of difference in neighbourhood characteristics between Simple Pearson and Artificial Immune System algorithms. Four characteristics are shown. In each case, the Wilcoxon significance test was applied to the neighbourhoods obtained from the algorithms. In all four cases, the value for the Simple Pearson was significantly higher; this is indicated by **bold** type.

| 1st Predictor | 2nd Predictor | Neighbourhood characteristic tested | Mean 1 | Mean 2 | Number of (unequal) neighbourhoods compared | 1st neighbourhood has higher value (sum of ranks) | 2nd neighbourhood has higher value (sum of ranks) | Significance (upper bound) |
|---|---|---|---|---|---|---|---|---|
| **Simple Pearson** | Artificial Immune System | Neighbours | 95.40 | 73.75 | 97 | 4602 | 151 | 1.196e-15 |
| **Simple Pearson** | Artificial Immune System | Overlap | 47.46 | 46.39 | 26 | 334.50 | 16.50 | 5.686e-05 |
| **Simple Pearson** | Artificial Immune System | Correlation | 0.12 | 0.10 | 79 | 2566 | 594 | 1.465e-06 |
| **Simple Pearson** | Artificial Immune System | Neighbour correlation | 0.15 | 0.04 | 83 | 3477 | 9 | 3.572e-15 |

**Fig 5a**

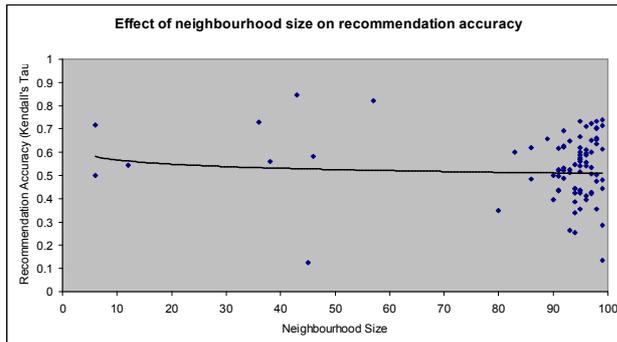

**Fig 5b**

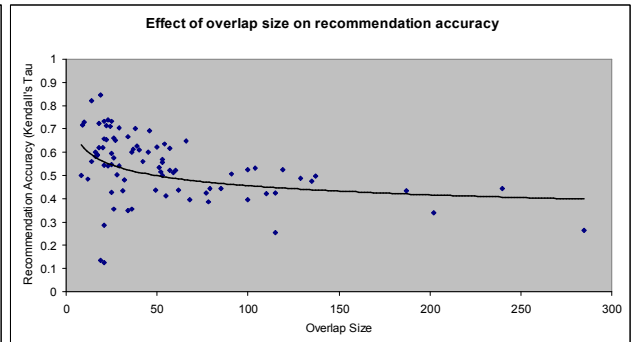

**Fig 5c**

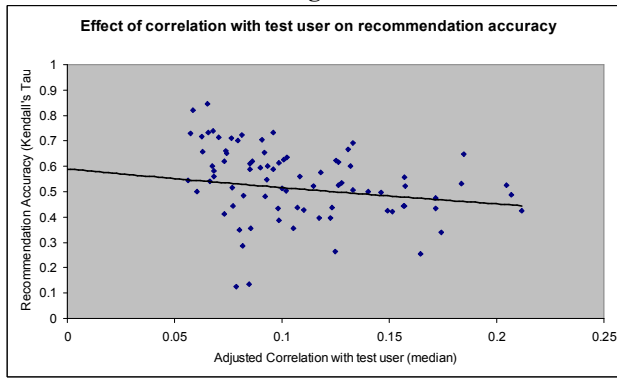

**Fig 5d**

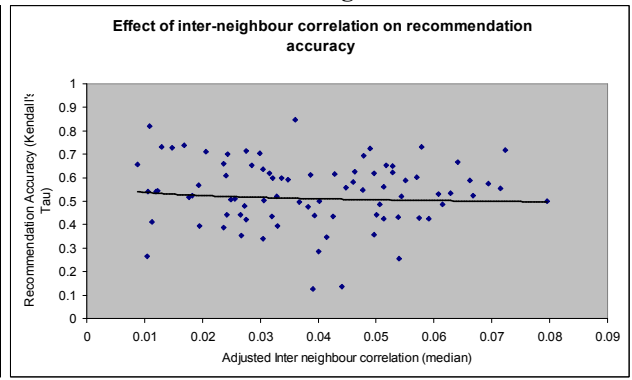

Figure 5. Effect of various neighbourhood measures on Artificial Immune System recommendation performance. In each graph, the measure is shown on the x-axis. The recommendation performance (where available) for each of 100 Artificial Immune System predictions is plotted against this neighbourhood measure. Trend lines are added to indicate the underlying data trend (if any).

## 5 DISCUSSION AND CONCLUSIONS

As mentioned previously, it is not claimed that these results are conclusive. Indeed, much more data is required before any firm conclusions can be drawn. In this respect, this paper is very much a work in progress. Nevertheless, the results to date certainly are indicative, and challenge certain assumptions. It is hoped that the presentation of these results will stimulate discussion and interest in the nature of the idiotypic effect.

It does not seem likely that the idiotypic effect can be captured by one particular measurement. Nevertheless, it is likely to be some combination of factors. For example, we have shown that both the neighbourhood choice and the weighting of neighbours within that neighbourhood can influence the recommendation performance. Pinning down the effect further has proved to be problematic. Our first intuition – that spreading out neighbours by reducing inter-neighbour correlation improves recommendation – appears to be at best incomplete and at worst incorrect. The mechanisms underlying the effect are clearly subtler than this.

There are of course other community characteristics that we could explore. Some (for example, number of recommendations, overlaps per neighbour, absolute correlation scores) have been examined and shown to be equally inconclusive. Some (for example, number of neighbours voting on each film) remain potential future subjects for investigation.

Other tests (e.g. setting each neighbour's concentration to a random number for immune system predictions, to see whether accurate concentrations are really necessary) might shed further light on the relative importance of each measure. But it is our intuition that such studies might not really get at the nature of the effect, and that larger scale or more sophisticated tests will be needed, coupled with perhaps analytical work, to get at the heart of this intriguing phenomenon.

There are wider implications for such work. The database used for this study [2] is based on real peoples' profiles. Thus, any headway made into improving neighbourhoods by the idiotypic effect can have real benefit for other recommenders – and indeed any community based application.

### References

[1] Cayzer S, Aickelin U, A Recommender System based on the Immune Network, Proceedings of the 2002 Congress on Evolutionary Computation, 2002.

[2] Compaq Systems Research Centre. EachMovie collaborative filtering data set, http://www.research.compaq.com/SRC/eachmovie/.

[3] Farmer JD, Packard NH and Perelson AS, The immune system, adaptation, and machine learning Physica, vol. 22, pp. 187-204, 1986.

[4] Gokhale A, Improvements to Collaborative Filtering Algorithms 1999. Worcester Polytechnic Institute. http://www.cs.wpi.edu/~claypool/ms/cf-improve/.

[5] Goldsby R, Kindt T, Osborne B, Kuby Immunology, Fourth Edition, W H Freeman, 2000.

[6] Jerne NK, Towards a network theory of the immune system Annals of Immunology, vol. 125, no. C, pp. 373-389, 1973.

[7] Perelson AS and Weisbuch G, Immunology for physicists Reviews of Modern Physics, vol. 69, pp. 1219-1267, 1997.